\documentclass[letterpaper]{article} 
\usepackage{aaai24}  
\usepackage{times}  
\usepackage{helvet}  
\usepackage{courier}  
\usepackage[hyphens]{url}  
\usepackage{graphicx} 
\usepackage{natbib}  
\usepackage{caption} 
\usepackage{algorithm}
\usepackage{algorithmic}

\usepackage{newfloat}
\usepackage{listings}
\DeclareCaptionStyle{ruled}{labelfont=normalfont,labelsep=colon,strut=off} 
\floatstyle{ruled}
\newfloat{listing}{tb}{lst}{}
\floatname{listing}{Listing}
\title{Geometric-Facilitated Denoising Diffusion Model for 3D Molecule Generation}
\author {
    Can Xu\textsuperscript{\rm 1,\rm 2}\footnote{This work is done during internship at Zhejiang Lab.}\footnote{Equal contribution, co-first author.},
    Haosen Wang\textsuperscript{\rm 3,\rm 2}\footnotemark[1],
    Weigang Wang\textsuperscript{\rm 1}\footnotemark[2],
    Pengfei Zheng\textsuperscript{\rm 2},
    Hongyang Chen\textsuperscript{\rm 2}\footnote{Correspond to Hongyang Chen.}
}
\affiliations {
    \textsuperscript{\rm 1}Zhejiang Gongshang University\\
    \textsuperscript{\rm 2}Zhejiang Lab\\
    \textsuperscript{\rm 3}Southeast University\\
    leoxc1571@163.com, haosenwang@seu.edu.cn, wangweigang@zjgsu.edu.cn, zpf2021@zhejianglab.com, dr.h.chen@ieee.org
}

\usepackage{makecell}
\usepackage{multirow}
\usepackage{amsfonts}

\begin{document}

\maketitle

\begin{abstract}
Denoising diffusion models have shown great potential in multiple research areas. Existing diffusion-based generative methods on {\itshape de novo} 3D molecule generation face two major challenges. Since majority heavy atoms in molecules allow connections to multiple atoms through single bonds, solely using pair-wise distance to model molecule geometries is insufficient. Therefore, the first one involves proposing an effective neural network as the denoising kernel that is capable to capture complex multi-body interatomic relationships and learn high-quality features. Due to the discrete nature of graphs, mainstream diffusion-based methods for molecules heavily rely on predefined rules and generate edges in an indirect manner. The second challenge involves accommodating molecule generation to diffusion and accurately predicting the existence of bonds. In our research, we view the iterative way of updating molecule conformations in diffusion process is consistent with molecular dynamics and introduce a novel molecule generation method named Geometric-Facilitated Molecular Diffusion (GFMDiff). For the first challenge, we introduce a Dual-Track Transformer Network (DTN) to fully excevate global spatial relationships and learn high quality representations which contribute to accurate predictions of features and geometries. As for the second challenge, we design Geometric-Facilitated Loss (GFLoss) which intervenes the formation of bonds during the training period, instead of directly embedding edges into the latent space. Comprehensive experiments on current benchmarks demonstrate the superiority of GFMDiff.
\end{abstract}

\section{Introduction}
Recent advances in deep generative methods, especially diffusion-based methods \cite{ddpm_20_ho,sgm_19_song,scoresde_21_song}, have greatly propelled research in generative artificial intelligence across various domains. In line with the development trend of generative methods, mainstream approches in the field of molecule discovery have undergone a transformation from previous generative methods to diffusion-based methods, and from designing 2D graphs to 3D conformations. However, there are two major challenges in 3D molecule generation. The first one involves predicting accurate and stable molecule conformations while the other entails fully utilizing geometric information to facilitate the generation of discrete graph structures. In this paper, we propose \textbf{Geometric-Facilitated Molecular Diffusion} (GFMDiff), a {\itshape de novo} 3D molecule generative method that addresses the aforementioned challenges. GFMDiff is capable of generating accurate 3D geometries while tackling the dicrete nature of graphs.


\begin{figure*}[t]
    \centering
    \includegraphics[width=0.8\linewidth]{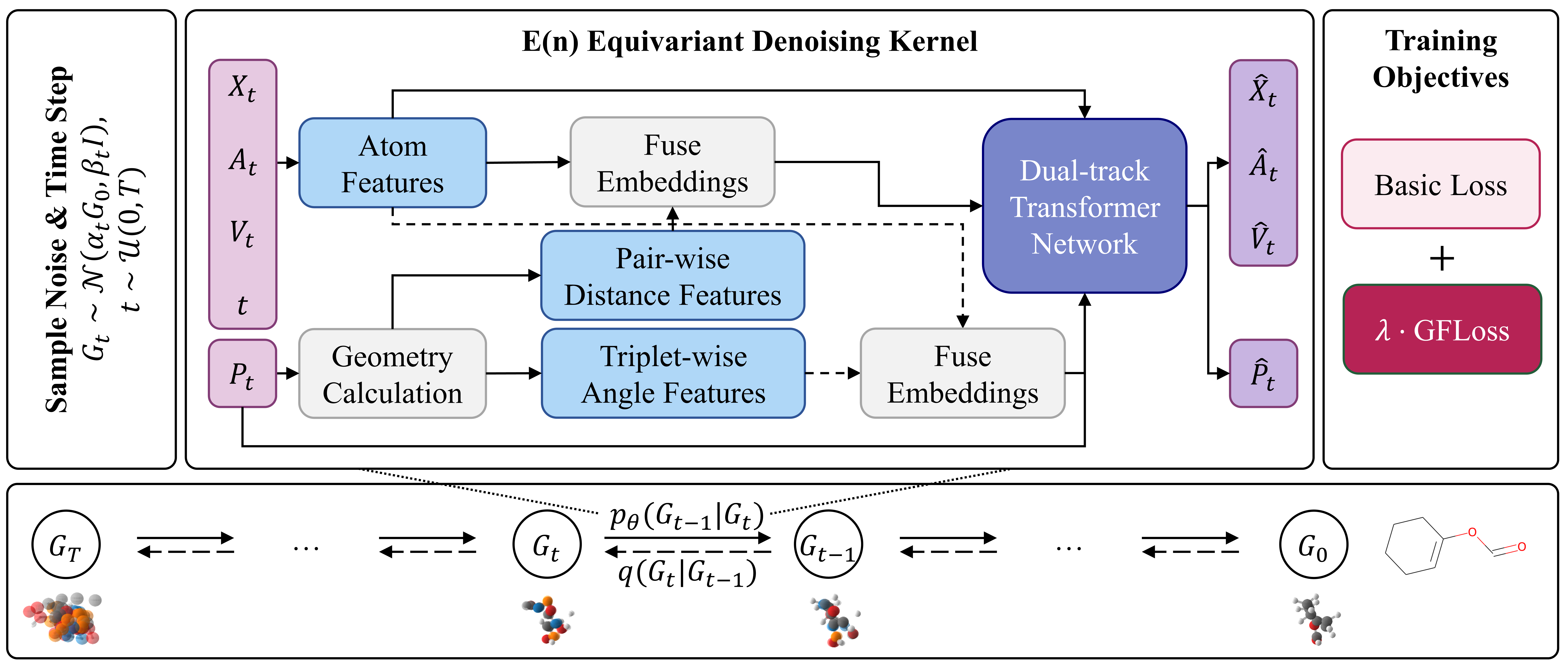}
    \caption{An overview of GFMDiff. For each noise sample at arbitrary time step, the denoising kernel predicts atom types, valencies, and coordinates through the DTN. A loss term named GFLoss that intervenes in the formation of bonds is added to the training objective as well.}
    \label{fig:gfmdiff}
  \end{figure*}

{\itshape De novo} molecule generation involves generating valid, novel, and stable molecules. To address the quest for equivariance of generated 3D conformations, several diffusion-based methods \cite{edm_22_hoogeboom,mdm_23_huang} model molecules indirectly through atomic distances, which directly reflet the strength of interatomic forces. However, earlier methods \cite{edm_22_hoogeboom} did not address the complex interatomic relationships among multiple atoms. While the others \cite{mdm_23_huang} simply use a threshold to distinguish the influence caused by chemical bonds and interatomic forces, regradlesss of specific atom and bond types. However, apart from few atoms like fluorine, chlorine, and bromine, majority heavy atoms in molecules allows connecting to other atoms through single bonds. Therefore, we consider simply using pair-wise distances to model molecule geometries is far from sufficient. Recent research \cite{moleformer_23_yuan} also indicates that the contribution of bond angles on molecular learning is equivalent to pair-wise distances. Only a limited number of methods make full use of spatial information. Additionally, since molecular diffusion methods only act on points cloud, traditional graph convolutions are unable to discriminate the importance of different atoms. In light of these challenges, we design a novel dual track molecular learning framework named \textbf{Dual-Track Transformer Network} (DTN). By integrating global transformer architecture, DTN serves as the denoising kernel to comprehensively capture spatial geometric information. 

Given the excellent performance of diffusion models on continuous data, majority of the models for molecule graph generation adopt the diffusion and denoising approach on Cartesian coordinates and features, followed by graph generation based on predefined rules instead of directly predicting the existence of bonds. The manner in which graphs are generated indirectly can potentially lead to degradation in stabilities and validity of generated samples. To make diffusion model applicable to multimodal data of molecules, several studies \cite{edpgnn_20_niu,digress_22_vignac,midi_23_vignac} introduce adjancency matrices to diffusion and denoising process. However, embedding graphs and edges into models leads to the rise of computational cost. In our research, we view diffusion model as a process that iteratively update atom information according to local multi-body interatomic relationship at each time step. Accurate feature learning assists on precise predictions of molecule conformaitons. In order to predict spot on molecule conformations, we devise a way by mitigating the gap between embeddings and local geometries during training. In this paper, we actively intervene the formation of graphs with a delicately designed loss funtion named \textbf{Geometric-Facilitated Loss} (GFLoss). 


In this paper, we present \textbf{Geometric-Facilitated Molecular Diffusion} (GFMDiff) for 3D molecule generation. In contrast to previous methods that primarily learn atom representations based-on pair-wise distances, we manage to effectively incorporate triplet-wise geometric information along with pair-wise distances into molecular learning. Most studies directly generate point clouds and subsequently complete 3D graph structures based on predefined rules. However, this approach suffers from two major constraints. Firstly, the indirect manner of graph generation causes degradation of stability and validity of samples. Secondly, traditional graph convolutions are not sufficient enough to distinguish local and global information. To address the first constraint, we proactively guide the formation of bonds during the training phase with a exquisite loss function named \textbf{Geometric-Facilitated Loss} (GFLoss). As for the second constraint, we introduce \textbf{Dual-Track Transformer Network} (DTN), a global transformer-based neural network, to promote comprehensive geometric learning and local feature learning. Main contributions of this paper are as follows:
\begin{itemize}
    \item Comprehensive utilization of spatial information to capture multi-body interactions among atoms, which is crucial for molecular learning and stabilities of generated samples.
    \item Introduction of a carefully designed GFLoss to facilitate the formation of bonds, addressing the discrete nature of graphs in an efficient manner.
    \item Proposal of DTN as an alternative to global graph convolutions which enables the model to capture both global and local information effectively.
\end{itemize}

\begin{figure*}[t]
    \centering
    \includegraphics[width=0.8\linewidth]{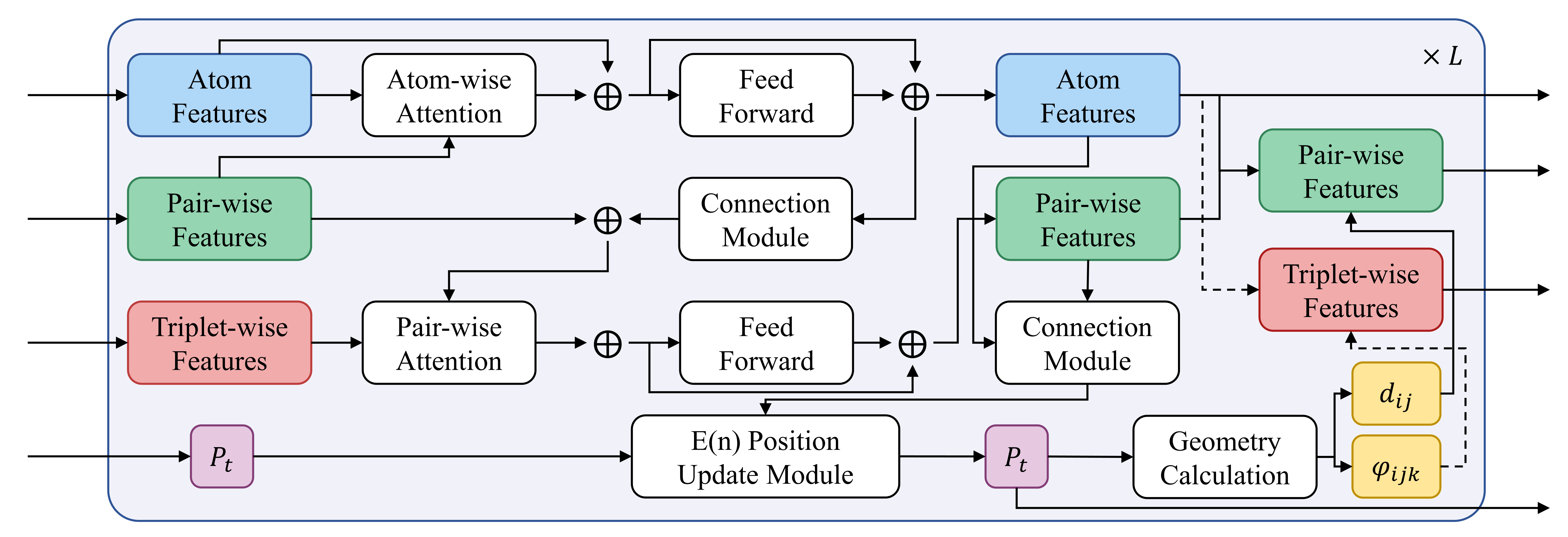}
    \caption{The illustration of Dual-Track Transformer Network (DTN). The atom-pair track and pair-triplet with multi-head attention modules update atom and pair-wise features, respectively. The pair-wise and triplet-wise features are further fused with the latest position information.}
    \label{fig:dtn}
\end{figure*}

\section{Related Works}
\subsection{Diffusion Models}
Diffusion-based methods arouse wide attention due to their excellent performance in generative tasks across multiple fields, such as computer vision \cite{blendeddiffusion_22_avrahami,cascadeddiff_22_ho,gradforshapegen_20_cai,sgmpointcloud_21_luo}, natural language processing \cite{argmaxflow_21_hoogeboom,stepunrolled_22_savinov}, as well as various interdisciplinary tasks \cite{housediffusion_23_shabani,nap_23_lei} et al. 

The formation of human motion skeletons share similarities to molecules, as both are represented by point clouds connected by edges. The difference is that human motion generation does not require predictions of edges. Building upon DDPMs, MoFusion \cite{mofusion_22_dabral} employs U-Net \cite{unet_15_ronneberger} as the backbone for the denosing kernel in motion sequence synthesis. 
Apart from applications on continuous data, many research efforts are devoted to discete data generation. EDP-GNN \cite{edpgnn_20_niu} introduce Score SDEs, a learning paradigm of diffusion models, to the generation of discrete graph adjancency matrices.

\subsection{Molecule Generation}
Over the past years, various generative methods have been employed for molecule generation, including variational autoencoders (VAEs) \cite{vae_13_kingma}, normalizing flows (NFs) \cite{nice_15_dinh}, and generative adversarial network (GANs) \cite{gan_14_goodfellow}. In order to generate 3D molecules, G-SchNet \cite{gschnet_19_wallach} adopts an auto-regressive model with a rotation invariant and local symmetrical network to add atoms incrementally. 
In the field of computatinal chemistry, diffusion models, well-suited for continuous data, are first introduced to conformation generation that takes molecule graphs as input and only operates on atom coordinates. 
EDM \cite{edm_22_hoogeboom} and MDM \cite{mdm_23_huang} convert discrete atom features into one-hot format to make generated samples more chemical feasible and stable. In order to generate edges without relying on predefined rules, Digress \cite{digress_22_vignac} and MiDi \cite{midi_23_vignac} propose discrete diffusion techniques. 



\section{Preliminaries}
\subsection{Denoising Diffusion Probabilistic Models}
Diffusion models have garnered considerable attention as powerful generative models. By learning the denoising kernel of the reverse process, these models are able to uncover underlying distributions of noisy samples. Given a piece of data, the forward process, treated as a Markov chain, gradually adds Gaussian noises for $T$ times to with learnable parameters controlling the strength of noises. During the generative process, the model reverses the noise back to the original distribution of real data.

\subsubsection{Diffusion Process}
Let $G_t (t=0, 1, ..., T)$ denotes distributions of molecule geometry information and $\beta_t \in (0, 1), t=0, 1, ..., T$ denote the variance schedule of the Markov chain. Therefore we have the posterior distribution of $G_t$:
\begin{eqnarray}
  &q(G_{1:T} | G_0) = \prod^T_{t=1} q(G_t | G_{t-1}), &\\ 
  &q(G_t | G_{t-1}) = \mathcal{N}(G_t; \sqrt{1-\beta_t}G_{t-1}, \beta_t I).&
\end{eqnarray}

As time step $t$ rises, the variance schedule $\beta_t$ smoothly transits from 0 to 1, which means more noise is added to the original data distribution. Let $\bar{\alpha}_t = \prod^t_{s=1} \alpha_s = \prod^t_{s=1}(1-\beta_s)$, the distribution of sample data
\begin{equation}
    q(G_t|G_0) = \mathcal{N}(G_t; \sqrt{\bar{\alpha}_t} G_0, (1 - \bar{\alpha}_t) I).
\end{equation} 

\subsubsection{Denoising Process}
During the denoising process, the model manages to reconstruct the original geometry information by learning Markov kernels $p_\theta(G_{0:T-1}| G_{T}) = \prod^T_{t=1} p_\theta(G_{t-1} | G_t)$ close to the actual reverse process $q(G_{t-1} | G_t)$. The distribution of learned parameterized dynamics at each time step is: 
\begin{equation}
  p_\theta(G_{t-1} | G_t) = \mathcal{N}(G_{t-1}; \mu_\theta(G_t, t), \sigma_t^2 I),
\end{equation}
where $\mu_\theta(G_t, t)$ is the neural network to approximate the means and $\sigma^2_t = \frac{(\beta_t - \beta_{t-1})\beta_{t-1}}{(1 - \beta_{t-1}) \beta_t}$ is the predefined variance schedule. Initially, the $p_\theta(G_t)$ is sampled from a standard Guassian distribution. Then the geometry information and atom features get polished over the reverse process iteratively. 

Theoretically, the training objective takes the form of the variational lower bound of log-likelihood of data:
\begin{eqnarray}
    &{\rm log}p(G) \geq \mathcal{L}_{base} + \sum_{t=0}^T \mathcal{L}_t,& \\
    &\mathcal{L}_{base} = -KL(q(G_T|G_0)|p(G_T)),& \\
    &\mathcal{L}_t = KL(q(G_{t-1}|G_t)|p(G_{t-1}|G_t)). &
\end{eqnarray}
However, it is found out that predicting the Guassian noise $\epsilon$ makes it easier for the neural network training. Therefore, $\mathcal{L}_t$ \cite{vaediff_21_kingma} takes the form of
\begin{equation}
    \mathcal{L}_t = E_{\epsilon_t \sim \mathcal{N}(0, I)}\left(\frac{1}{2}(1- \frac{{\rm SNR}(t-1)}{{\rm SNR}(t)})||\epsilon_t - \hat{\epsilon}_t||^2\right).
\end{equation}


\section{Methodology}
In this section, we provide details of our proposed molecule generation framework, including the E(n) equivariant denoising kernel, Geometric-Facilitated loss, the forward and reverse process, and the optimization objective. The overview of GFMDiff is shown in Fig~\ref{fig:gfmdiff}. At each time step, molecule conformations are sampled as inputs. The DTN ensures complete utilization of molecule geometries by taking pair-wise distances and triplet-wise angles as inputs. These two kinds of features each stand for interatomic forces and multi-body interactions. The incorporation of GFLoss further guarantees the reasonableness and soundness of generated samples.

\subsection{Dual-Track Transformer Network}
In this subseciton, we elaborate on \textbf{Dual-Track Transformer Network} (DTN) as the E(n) equivariant backbone of GFMDiff. DTN is designed to effectively capture interatomic relationships and atom features. Since 3D  molecular geometries possess invariance properties such as rotations, translations, reflections, and permutations, it is essential for the denoising kernel to satisfy such properties. The proposed DTN is not only E(n) equivariant, but also able to fully leverages spatial information.


In our proposed method, we regard an input molecule with the total number of atoms $N$ as $G = (P, X, A, V)$, where $P = (p_1, p_2, ..., p_N) \in \mathbb{R}^{N \times 3}$ represents atom coordinates, $X = (x_1, x_2, ..., x_N) \in \mathbb{R}^{N \times nf}$ represents one-hot encoding of atomic numbers, $A = (a_1, a_2, ..., a_N) \in \mathbb{R}^{N}$ represents atomic numbers, and $V = (v_1, v_2, ..., v_N) \in \mathbb{R}^{N}$ stands for valencies of atoms. To ensure the quest for the equivariance, DTN utilizes pair-wise distances and triplet-wise angles to capture the geometry information. The Euclidean distance between atom $i$ and $j$, which reflects the strength of interatomic force, is obtained by:
\begin{equation}
    d_{ij} = ||p_i - p_j||_2.
\end{equation}
Given the complicated relationships among atoms during the sampling phase, simply using pair-wise distances is insufficient to excevate spatial information. Therefore, we further calculate the triplet-wise angle using:
\begin{equation}
    \varphi_{ijk} = \arccos \left( \frac{(p_i - p_j) \times (p_i - p_k)}{||p_i - p_j||_2 \times ||p_i - p_k||_2} \right). 
\end{equation}
The above geometric calculations allow us to further featurize local geometries through radius basis function (RBF) network:
\begin{eqnarray}
    &e_{ij} = {\rm Linear}({\rm RBF}(d_{ij}), e_i, e_j),& \\
    &e_{ijk} = {\rm Linear}({\rm RBF}(\varphi_{ijk}), e_i, e_j, e_k),&
\end{eqnarray}
where $e_i = {\rm Embedding}(x_i, a_i, v_i)$ is the node embedding of atom $i$ that combines atomic numbers and valencies. These features are later fed into DTN with $L$ layers. 

\begin{figure}[t]
    \centering
    \includegraphics[width=0.8\linewidth]{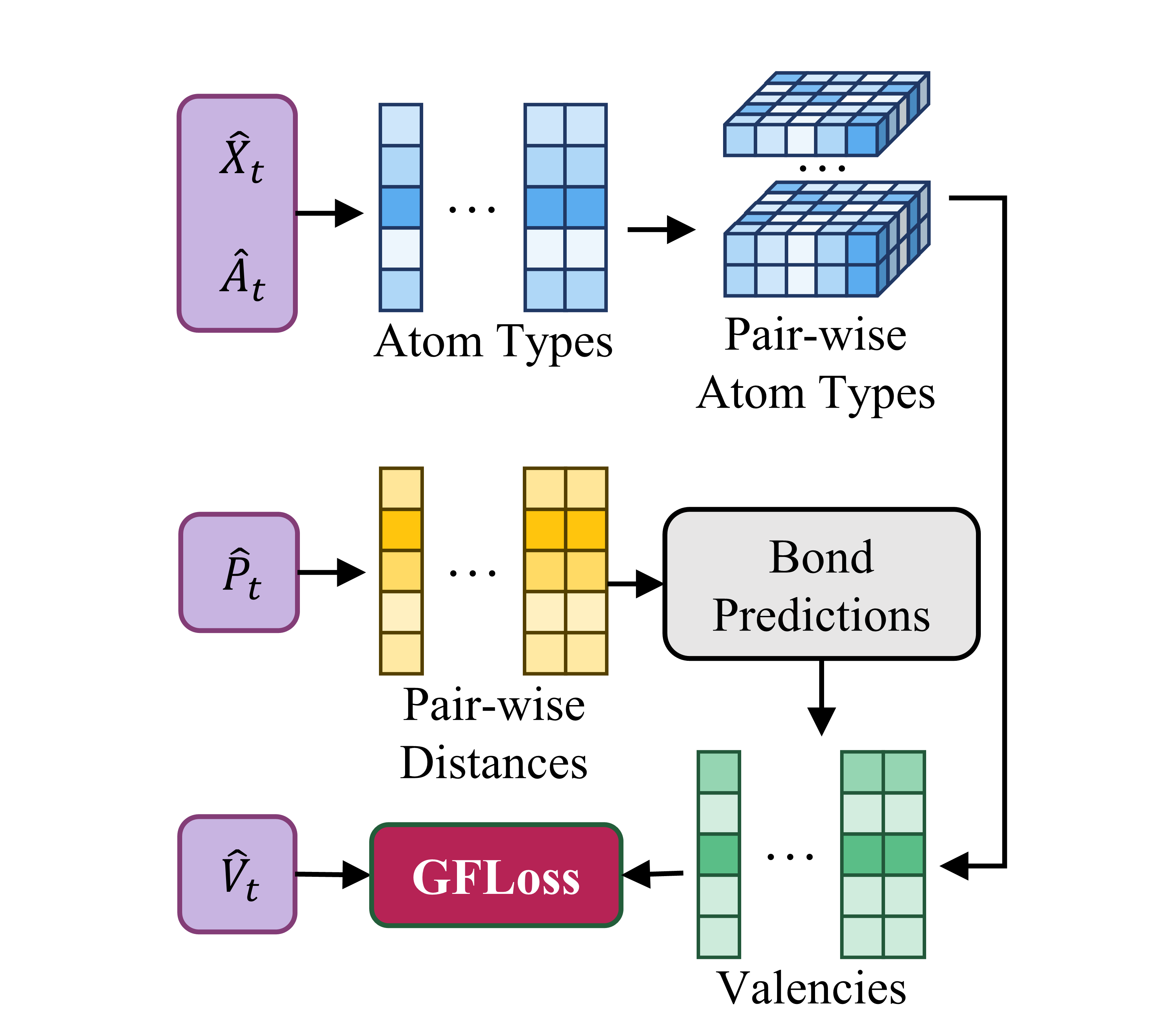}
    \caption{The illustration of GFLoss. We leverages chemical rules to predict the existence of bonds and then calculate potential valencies based on probabilities of atom types and bond predictions. The loss minimizes the difference between valencies predicted by DTN and valencies calculated from molecule geometries.}
    \label{fig:gfloss}
\end{figure}

Each layer of DTN consists of the following components, an atom-pair track, a pair-triplet track, and a connection module. The atom-pair track simulates the influence of interatomic forces on atoms while the pair-triplet track models the impact of potential bond angles on edges. Just as its name implies, the connection module serves at the brige between two tracks to promote better representation learning by injecting atom features back to pair-wise features.

The atom-pair track involves predicting influences of other atoms and interatomic forces on target atoms. The track takes atom embeddings $e_i$ and pair embeddings $e_{ij}$ as inputs: 
\begin{eqnarray}
    &e_i = {\rm LayerNorm}(e_i), \ e_{ij} = {\rm LayerNorm}(e_{ij}),& \\
    &\mathbf{Q}_i = {\rm Linear}(e_i), \ \mathbf{K}_i = {\rm Linear}(e_i) + {\rm Linear}(e_{ij}),& \\
    &a_i = {\rm Dropout}({\rm softmax} \frac{\mathbf{Q}_i \mathbf{K}_i^T}{\sqrt{d_h}}), & \\
    &\mathbf{V}_i = {\rm Linear}(e_{ij}) + {\rm Linear}(e_{i}) + {\rm Linear}(e_{j}),& \\
    &\hat{e}_i = {\rm Linear}(a_i \mathbf{V}_i^T),&
\end{eqnarray}
where $d_h$ is the number of heads. Atom embeddings first get updated by adding predictions of atom-pair track and are later passed to a feed forward network. In each layer, atom absorbs aggregated representations of other atoms and corresponding atom pairs.

Similarly, the pair-triplet track predicts the impact of complex geometry substructures on interatomic forces.
It is worth noting that triplet embeddings $e_{ijk}$ does not get updated in the transformer structure, because this would significantly increase the quest for computaional resources. They only get updated whenever atom coordinates are renewed.

The connection module serves as the role that fuse atom embeddings into pair embeddings. For pair embedding $e_{ij}$, it absorbs atomic feature information from the connection module and local spatial information from the pair-triplet track at the same time.
\begin{equation}
    e_{ij} = {\rm LayerNorm}\left(e_{ij} + {\rm Linear}({\rm Linear}(e_i) \otimes {\rm Linear}(e_j))\right)
\end{equation}

In terms of the approach to update coordinates, we follow the design of PosUpdate module in EDM \cite{edm_22_hoogeboom} and MDM \cite{mdm_23_huang}. At the end of each layer, pair-wise and triplet-wise embeddings get updated since molecule conformations are altered:
\begin{eqnarray}
    &e_{ij} = {\rm Linear}\left({\rm Linear}({\rm RBF}(\hat{d}_{ij}), \hat{e}_{ij}), \hat{e}_i, \hat{e_j}\right),&\\
    &e_{ijk} = {\rm Linear}\left({\rm RBF}(\hat{\varphi}_{ijk}), e_{ijk}\right). &
\end{eqnarray}

\begin{table*}[t]
    \centering
    \begin{tabular}{lllllll}
    \hline
    Type & Method & NLL$\downarrow$ & \makecell[l]{Atom\\Stability (\%)} $\uparrow$ & \makecell[l]{Mol\\Stability (\%)} $\uparrow$ & Validity (\%) $\uparrow$ & Uniqueness$\cdot$Validity (\%) $\uparrow$ \\
    \hline
    NF & E-NF & -59.7 & 85.0 & 4.9 & 40.2 & 39.4 \\
    AR & G-SchNet & N/A & 95.7 & 68.1 & 85.5 & 80.3 \\
    \multirow{2}{*}{DDPM}
    & EDM & -110.7$\pm$1.5 & 98.7$\pm$0.1 & 82.0$\pm$0.4 & 91.9$\pm$0.5 & 90.7$\pm$0.6 \\
    & Bridge+Force & N/A & 98.8$\pm$0.1 & 84.6$\pm$0.3 & N/A & N/A \\
    & GCDM & \textbf{-171.0}$\pm$0.2 & 98.7$\pm$0.0 & 85.7$\pm$0.4 & 94.8$\pm$0.2 & 93.3$\pm$0.0 \\
    & GeoLDM & N/A & \underline{98.9}$\pm$0.1 & \textbf{89.4}$\pm$0.5 & 93.8$\pm$0.4 & 92.7$\pm$0.5 \\
    \hline
    \multirow{2}{*}{Ours} & \makecell[l]{GFMDiff\\w/o tri} & -123.1$\pm$0.4 & 98.7$\pm$0.1 & 85.9$\pm$0.2 & 94.9$\pm$0.2 & 94.2$\pm$0.2 \\
    & \makecell[l]{GFMDiff\\w/o GFLoss} & -127.5$\pm$0.4 & 98.7$\pm$0.0 & 86.5$\pm$0.1 & \underline{95.2}$\pm$0.0 & \underline{94.5}$\pm$0.0 \\
    & GFMDiff & \underline{-128.0}$\pm$0.2 & \textbf{98.9}$\pm$0.0 & \underline{87.7}$\pm$0.2 & \textbf{96.3}$\pm$0.3 & \textbf{95.1}$\pm$0.2 \\
    \hline
    Data &  &  & 99.0 & 95.2 & 97.7 & 97.7 \\
    \hline
    \end{tabular}
    \caption{Performance comparison on GEOM-QM9. Results of 10000 generated samples are reported with standard deviations across 3 runs using different seeds.}
    \label{tab:exp_qm9}
\end{table*}
\begin{figure*}[t]
    \centering
    \includegraphics[width=\linewidth]{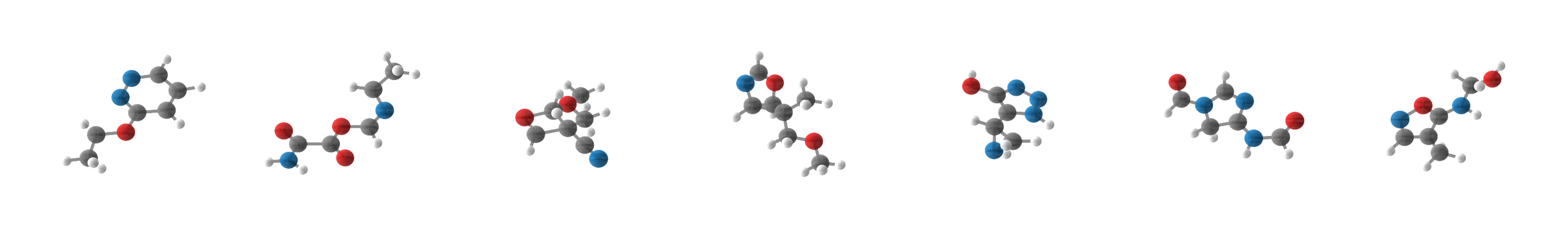}
    \caption{Molecule samples generated by GFMDiff for GEOM-QM9}
    \label{fig:samples_qm9}
\end{figure*}

\begin{table*}[t]
    \centering
    \begin{tabular}{lllllll}
    \hline
    \makecell[l]{Task\\Units} & \makecell[l]{$\alpha$\\${\rm Bohr^3}$} & \makecell[l]{$\Delta \varepsilon$\\${\rm meV}$} & \makecell[l]{$\varepsilon_{{\rm HOMO}}$\\${\rm meV}$} & \makecell[l]{$\varepsilon_{{\rm LUMO}}$\\${\rm meV}$} & \makecell[l]{$\mu$\\${\rm D}$} & \makecell[l]{$C_v$\\$\frac{{\rm cal}}{{\rm mol}}{\rm K}$} \\
    \hline
    Naive (Upper-bound) & 9.01 & 1470 & 645 & 1457 & 1.616 & 6.857 \\
    \# Atom & 3.86 & 866 & 426 & 813 & 1.053 & 1.971 \\
    EDM & 2.76 & 655 & 356 & 584 & 1.111 & 1.101 \\
    GCDM & \underline{1.97} & 602 & 344 & \underline{479} & \underline{0.844} & \underline{0.689} \\
    GeoLDM & 2.37 & \underline{587} & \underline{340} & 522 & 1.108 & 1.025 \\
    GFMDiff & \textbf{1.74} & \textbf{558} & \textbf{321} & \textbf{430} & \textbf{0.728} & \textbf{0.593} \\
    QM9 (Lower-bound) & 0.10 & 64 & 39 & 36 & 0.043 & 0.040 \\
    \hline
    \end{tabular}
    \caption{Performance comparison for conditioned molecule generation on QM9. With conditioned samples, Results are in the form of mean absolute error (MAE) for property prediction of 10000 conditional samples by an EGNN classifier.}
    \label{tab:exp_qm9_cond}
\end{table*}
\begin{figure*}[t]
    \centering
    \includegraphics[width=\linewidth]{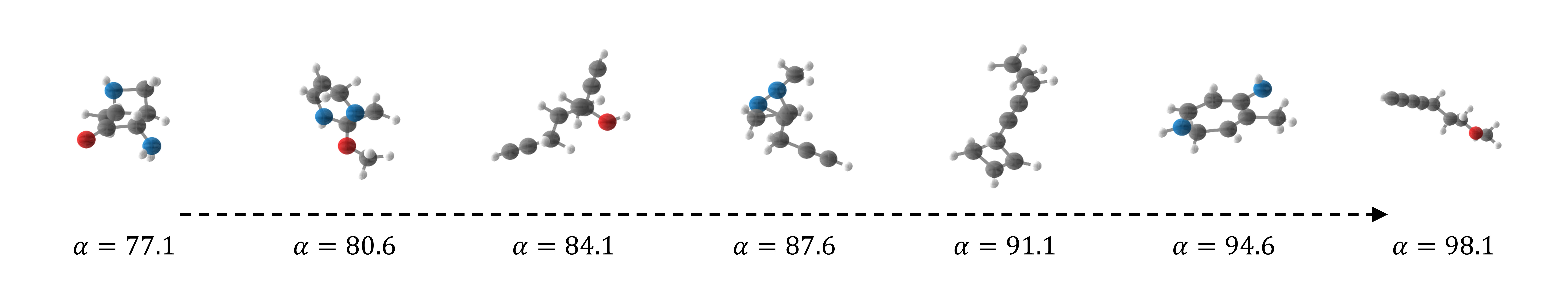}
    \caption{Generated samples of GFMDiff on QM9 conditioned with increasing values of $\alpha$}
    \label{fig:samples_qm9_cond}
\end{figure*}

\subsection{Geometric-Facilitated Loss}
Predicting the existence of bonds is a fundamental and indispensable task in molecule graph generation. Unlike previous reserch that heavily relies on predefined rules, we propose to actively intervened in bond formation during the training process by designing a delicate training objective term named \textbf{Geometric-Facilitated Loss} (GFLoss). The intention of this loss function is to guide the model in generating molecules that not only possess valid topological structures but also stable conformations. We consider the valencies of atoms to be the an type of auxiliary features of great importance in molecule generation. Therefore, valencies of atoms are incorporated as part of atom features in the aboved metioned DTN. The cross validation of valencies allows the model to establish close connections between geometries and validity.

According to the predefined rules, atom pairs with proper distances are considered to be connected by bonds. For single, double, or triple bond, there are typical distances between certain atoms. If the distance between a pair of atoms is in certain range, these two atoms are considered to be connect by corresponding type of bond. Let the predefined distances and margins to be $\mathbf{D} \in \mathbb{R}^{nf \times nf \times 3}$ and $\mathbf{M} \in \mathbb{R}^{3}$, where 3 stands for the number of bond types. Based on outputs of DTN $\hat{G}_t = (\hat{P}_t, \hat{X}_t, \hat{A}_t, \hat{V}_t)$, we first predict the probabilities of atom types using a softmax function:
\begin{equation}
    {\mathbf{p}}_t(\hat{X}_{{\rm atom}}) = {\rm softmax}(\hat{X}_t) \in \mathbb{R}^{N \times nf},
\end{equation}
where $\hat{X}_t$ here represents the predicted atom types in one-hot format with the dimension of $nf$. The probabilities of pair-wise atom types are
\begin{equation}
    {\mathbf{p}}_t(\hat{X}_{{\rm pair}}) = {\mathbf{p}}_t(\hat{X}_{{\rm atom}}) \cdot {\mathbf{p}}_t(\hat{X}_{{\rm atom}}) \in \mathbb{R}^{N \times N \times nf \times nf}.
\end{equation}

With the predicted atom coordinates $\hat{P}_t$, pair-wise distance matrix $\mathbf{d}_t \in \mathbb{R}^{N \times N}$ can be obtained and is expanded to $\mathbb{R}^{N \times N \times nf \times nf \times 3}$ for convenience. Then the margin $\mathbf{m}_t$ between pair-wsie distances and typical bond distancess is:
\begin{equation}
    \mathbf{m}_t = \mathbf{d}_t - (\mathbf{D} + \mathbf{M}) \in \mathbb{R}^{N \times N \times nf \times nf \times 3}.
\end{equation}
Take atom $i$ and $j$ for an example, suppose their chances to be Carbon are above zero, if any element in margin $\mathbf{m}_t(i,j,{\rm C},{\rm C},:) \in \mathbb{R}^{3}$ is below zero, it indicates the presence of a bond between atom $i$ and $j$. Specific type of bond is determined by the index of the minimum value in $\mathbf{m}_t(i,j,{\rm C},{\rm C},:)$. If $\arg\min (\mathbf{m}_t(i,j,{\rm C},{\rm C},:))$ is 1, then they are connected by a single bond. They are connected by a triple bond if $\arg\min (\mathbf{m}_t(i,j,{\rm C},{\rm C},:))$ happen to be 3. The boolean matrix that represents the existence of bonds is noted as ${\rm ISBOND}_t \in \mathbb{R}^{N \times N \times nf \times nf}$

Once we have the probabilities of pair-wise atom types and the existence of bonds, the probable valencies of atoms could be estimated as:
\begin{equation}
     \hat{V}_{{\rm pred}}(t) = {\rm sum}({\mathbf{p}}_t(\hat{X}_{{\rm pair}}) \odot {\rm ISBOND}_t) \in \mathbb{R}^{N}.
\end{equation}
Since input data are fused with different level of noises, GFLoss is fomulated as the mean square error between the predicted valencies $V_{{\rm pred}}$ and the ground-truth valencies $V$:
\begin{eqnarray}
    &\mathcal{L}_t = E_{\epsilon_t \sim \mathcal{N}(0, I)}\left(\frac{1}{2}\omega(t)(||\epsilon_t - \hat{\epsilon}_t||^2 + \lambda \mathcal{L}_{GF}(t))\right), & \\
    &\mathcal{L}_{GF}(t) = ||\alpha_t(\hat{V}_{pred}(t) - V_t)||^2, &
\end{eqnarray}
where $\alpha_t$ is the level of ground-truth data in a piece of noisy input in diffusion process and $\omega(t) = (1-{\rm SNR}(t)/{\rm SNR}(t-1))$.

\begin{table*}[t]
    \centering
    \begin{tabular}{llll}
    \hline
    Type & Method & Atom Stable (\%) $\uparrow$ & Mol Stable (\%) $\uparrow$ \\
    \hline
    Normalizing flow & E-NF & 75.0 & 0 \\
    \multirow{2}{*}{DDPM}
    & EDM  & 81.3 & 0.0 \\
    & Bridge+Force & 82.4$\pm$0.8 & 0.0 \\
    & GCDM & \underline{86.4}$\pm$0.2 & \underline{3.7}$\pm$0.3 \\
    & GeoLDM & 84.4 & 3.2 \\
    \hline
    Ours & GFMDiff & \textbf{86.5}$\pm$0.2 & \textbf{3.9}$\pm$0.2 \\
    \hline
    Data &  & 86.5 & 2.8 \\
    \hline
    \end{tabular}
    \caption{Performance comparison on GEOM-Drugs. Results of 10000 generated samples are reported with standard deviations across 3 runs using different seeds.}
    \label{tab:exp_drugs}
\end{table*}
\begin{figure*}[t]
    \centering
    \includegraphics[width=\linewidth]{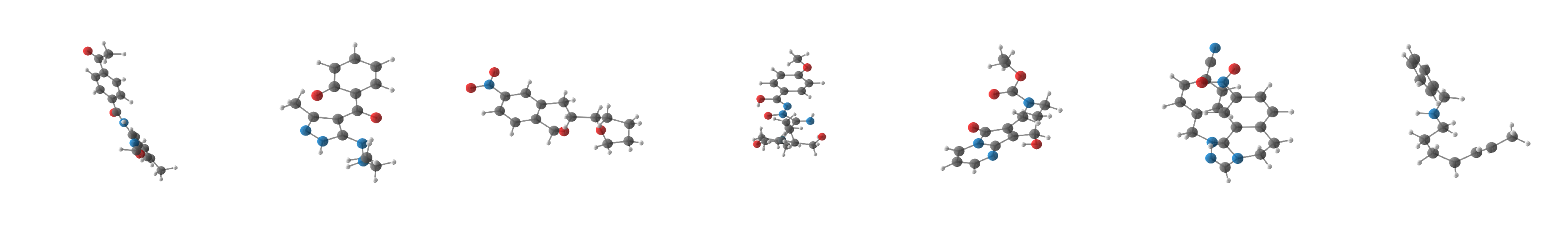}
    \caption{Molecule samples generated by GFMDiff for GEOM-Drugs}
    \label{fig:samples_drugs}
\end{figure*}

\section{Experiments}
In this section, we report the performance comparison of GFMDiff on GEOM-QM9 \cite{qm9_14_ramakrishnan} and GEOM-Drugs \cite{drugs_22_axelrod}. 
Results on three current benchmarks indicate that our method  outperforms state-of-the-art (SOTA) models in multiple aspacts.

\subsection{Setup}
In order to make comprehensive comparisons, we conduct experiments on two benchmark datasets in molecule generation: GEOM-QM9 \cite{qm9_14_ramakrishnan} and GEOM-Drugs \cite{drugs_22_axelrod}. GEOM-QM9 dataset consists of over 130K molecules and their corresponding conformations, where molecules have 18 atoms with hydrogen included on average. GEOM-Drugs includes over 450K molecules and 37M conformations, where size of molecule is 44 on average.

To assess the performance of GFMDiff in a fair and comprehensive manner, we compare it against six representative baselines in this fiels, which are E-NF \cite{enf_21_satorras}, G-SchNet \cite{gschnet_19_wallach}, EDM \cite{edm_22_hoogeboom}, models of Wu et al. \cite{diffpg_22_wu}, GCDM \cite{gcdm_23_morehead}, and GeoLDM \cite{geoldm_23_xu}. We refer to the performances of the first three models stated in EDM, as well as results of the remaining baselines reported in GCDM and GeoLDM. 

In terms of evaluation metrics, we adopt the same ones used in previous research, which are stability, validity, and uniqueness. Stability measures the proportion of atoms with correct valencies and molecules whose atoms are all stable. Validity is defined as the percentage of molecules that are theoretically correct and uniqueness shows the probability of non-repetitive samples. Arrows in Table~\ref{tab:exp_qm9} and Table~\ref{tab:exp_drugs} signify the preferred direction of each criteria. The best results are highlightened in bold and the second best results are underlined.

\subsection{{\itshape De Novo} Molecule Generation on QM9}
In order to analyze results fairly, we use the same dataset settings as previous methods. To evaluate the effectiveness of GFLoss and triplet geometric information learning, we includes GFMDiff w/o GFLoss and GFM w/o tri for comparison. In GFM w/o tri, we replace the pair-triplet track multi-head attention module with a self-attention module of pair-wise features. The weight for GFLoss $\lambda$ is set 0.01. On QM9, GFMDiff is trained for around 1000 epochs, with a five layer DTN and the embedding size of 256.

As it is shown in Table~\ref{tab:exp_qm9}, GFMDiff outperforms all baselines and achieves the best performance in stability, validity, and uniqueness times validity. GFMDiff and recent SOTA methods show no major difference in stability of atoms, but the performance lead of GFMDiff over the second-best method using the same generative methods in terms of stability of molecules is 2.1\%. This indicates that our model is capable of genrating stable molecules. We believe that the molecule stabilty could be further improved using latent diffusion in GeoLDM. The performance lead of GFMDiff over the SOTA method in validity and validity times uniqueness is 1.1\% and 1.3\%, respectively. The superior performance in validity means that GFMDiff generates molecules not only with accurate conformations, but also with correct valid and unique structres. It is intriguing to find out that GFMDiff exhibits lower performance in terms of the negative log-likelihood of data (NLL) compared to GCDM, but still surpasses other baselines. A possible explanation could be the different ways of applying geomteric information between GFMDiff and GCDM. 

Moreover, the ablations of GFLoss and triplet-wise geomtery illustrate the effectiveness of them. Among GFMDiff and its abalation models, GFMDiff w/o tri achieves the lowest results. This means the incorporation of complete local geometry information contributes more to the performance lift than GFLoss. In summary, GFMDiff exhibits the ability to generate stable molecules while addressing validities of samplessimultaneously.

\subsection{Conditional Molecule Generation on GEOM-QM9}
For conditional molecule generation on QM9, we compare our GFMDiff with existing methods along with naive baselines. In Table~\ref{tab:exp_qm9_cond}, we show the comparison of MAE on property prediction task. The "Naive (Upper-bound)" is a baseline where samples and labels are shuffled and the "\#Atoms" is the property prdiction method which simply relies of the number of atoms. Lower mean absolute errors of a model than these two baselines indicate the model is capable to incorporate properties and molecule conformation information into generated samples. 

As it is demonstrated in Table~\ref{tab:exp_qm9_cond}, our methods outperforms the state-of-the-art method in this task. Samples of samples with various values of $\alpha$ is shown in Figure~\ref{fig:samples_qm9_cond} as well. The performence lead of GFMDiff over the second-best method on for 6 properies are 11.7\%, 4.9\%, 6.2\%, 10.2\%, 13.7\%, and 13.9\%, respectively. Results indicates the superiority of our GFMDiff in generate molecules with desirable properties.

\subsection{{\itshape De Novo} Molecule Generation on GEOM-Drugs}
It is a challenging task to generate molecules for GEOM-Drugs dataset, since it is a large scale dataset of big molecules with up to 181 atoms. The relatively large scale of molecules and low stabilities of ground truth data bring huge challenges to 3D molecule generation. In experiments on GEOM-Drugs, we compares GFMDiff with E-NF, EDM, Bridge + Force, GCDM, and GeoLDM. Since current methods performs poorly in the novelty of molecules, we only list the stability of generated samples for comparison.

Due to the size of molecules in GEOM-Drugs, the stability of ground truth data are much lower than that in QM9. The proposed GFMDiff outperforms GCDM in terms of atom stability by a small margin, while GFMDiff outperforms the second-best result on molecule stability by 5.4\%. It's worth noting that GeoLDM, which generates samples with high stabilities on QM9, encounters a bottleneck in generating large molecules. Some samples generated by GFMDiff are shown in Figure~\ref{fig:samples_drugs}. Results on Drugs also demonstrate the capability of our proposed GFMDiff to generate stable molecule geometries.

\section{Conclusion}
In this paper, we propose GFMDiff, a novel molecule generation methods that fully excevates geometric information to help expressive representation learning and accurate bonds formation in molecule graphs. Unlike earlier methods that did not comprehensively model molecular geometries and heavily rely on predefined rules to generate bonds, GFMDiff makes full use of spatial information to assist on representation learning and facilitate accurate edge generation. We adopt DTN as the denoising kernel to update atom features and coordinates based on interatomic forces and multi-body interactions. The GFLoss is also implemented to actively intervene the formation of bonds during each time step at the training stage. We conduct comprehensive experiments to evaluate the effectiveness and performance edge of the proposed techniques over SOTA methods. It is shown that GFMDiff is capable to generate valid molecules with accurate conformations and correct atom valencies. 

\section{Acknowledgments}
This project is supported by National Key Research and Development Program of China (2022YFB4500300), the National Natural Science Foundation of China (72273132), in part by Key Research Project of Zhejiang Lab (No. 2022PI0AC01). We also gratefully acknowledge the valuable comments from anonymous reviewers.

\bibliography{bibfile_aaai24}

\end{document}